\pdfoutput=1

\documentclass[11pt]{article}
\usepackage[table]{xcolor}
\usepackage{array}

\usepackage{ifthen} 
\newboolean{review}
\setboolean{review}{false}  

\newcommand{\iffinal}[2]{%
  \ifthenelse{\boolean{review}}{#2}{#1}%
}
\newcommand{\ifrev}[2]{%
  \ifthenelse{\boolean{review}}{#1}{#2}%
}

\ifrev{\usepackage[review]{acl}}
    {\usepackage[preprint]{acl}}

\usepackage{lmodern}

\usepackage{times}
\usepackage{latexsym}

\usepackage[T1]{fontenc}

\usepackage[utf8]{inputenc}

\usepackage{microtype}

\usepackage{inconsolata}

\usepackage{graphicx}

\usepackage{hyperref}
\usepackage{url}
\usepackage{booktabs}
\usepackage{authblk}

\usepackage{enumitem}
\setlist{noitemsep}

\newcommand{\ttw}[1]{\textit{#1}}

\usepackage{multirow} 
\usepackage{nicematrix} 

\usepackage{relsize}
\usepackage{floatrow}
\DeclareFloatFont{tiny}{\small}
\usepackage{colortbl}
\usepackage{makecell}
\usepackage{pgfmath}

\definecolor{myblue}{rgb}{0.392, 0.584, 0.929}
\definecolor{mygreen}{rgb}{0.133, 0.545, 0.133}

\newcommand{\colg}[2]{%
  \cellcolor{mygreen!#1!white}\makecell[{{r}}]{#2}%
}
\newcommand{\colb}[2]{%
  \cellcolor{myblue!#1!white}\makecell[{{r}}]{#2}%
}

\newcommand{\sectionref}[1]{\hyperref[#1]{Section~\ref{#1}: \nameref{#1}}}
\newcommand{\sectionrefshort}[1]{\hyperref[#1]{§~\ref{#1}}}
\newcommand{\appref}[1]{\hyperref[#1]{Appendix~\ref{#1}}}
\newcommand{\apprefsh}[1]{\hyperref[#1]{App.~\ref{#1}}}
\newcommand{\apprefshort}[1]{\hyperref[#1]{\ref{#1}}}

\newcommand{\babylm}{BabyLM\xspace}
\newcommand{\babylms}{BabyLMs\xspace}
\newcommand{\babylmc}{BabyLM Challenge\xspace}

\usepackage{xspace}

\newcommand{\modelname}[1]{\texttt{#1}}
\newcommand{\mn}[1]{\modelname{#1}\xspace}

\newcommand{\ms}[2]{\mn{#1\textsubscript{#2}}}

\newcommand{\gptb}{\mn{GPT-BERT}}
\newcommand{\gptbb}{\ms{GPT-BERT}{100M}}
\newcommand{\gptbs}{\ms{GPT-BERT}{10M}}

\newcommand{\ltgb}{\mn{LTG-BERT}}
\newcommand{\ltgbb}{\ms{LTG-BERT}{100M}}
\newcommand{\ltgbs}{\ms{LTG-BERT}{10M}}

\newcommand{\btimeb}{\ms{BERTtime}{100M}}
\newcommand{\btimes}{\ms{BERTtime}{10M}}

\newcommand{\eli}{\ms{ELI5}{100M}}
\newcommand{\qecl}{\ms{QE CL}{10M}}

\newcommand{\robt}{RoBERTa\xspace}

\newcommand{\rob}{\mn{RoBERTa}}
\newcommand{\robl}{\ms{RoBERTa}{L}}
\newcommand{\robb}{\ms{RoBERTa}{B}}

\newcommand{\bbbb}{\mn{BERT}}
\newcommand{\bertl}{\ms{BERT}{L}}
\newcommand{\bertb}{\ms{BERT}{B}}

\title{
BabyLM's First Constructions: \\Causal probing provides a signal of learning
}

\renewcommand*{\Affilfont}{\normalsize\normalfont}

\makeatletter
\renewcommand\AB@affilsepx{\hspace{1em} \protect\Affilfont}
\makeatother

\author[1]{Joshua Rozner}
\author[2]{Leonie Weissweiler}
\author[1]{Cory Shain}

\affil[1]{Stanford University}
\affil[2]{Uppsala University\protect\\
\texttt{
rozner@stanford.edu,
leonie.weissweiler@lingfil.uu.se,
cory.shain@gmail.com
}}

\begin{document}
\maketitle
\begin{abstract}
Construction grammar posits that language learners acquire constructions (form-meaning pairings) from the statistics of their environment. 
Recent work supports this hypothesis by showing sensitivity to constructions in pretrained language models (PLMs), including one recent study \cite{rozner2025} demonstrating that constructions shape \rob's output distribution. 
However, models under study have generally been trained on developmentally implausible amounts of data, casting doubt on their relevance to human language learning. 
Here we use \citeauthor{rozner2025}'s methods to evaluate construction learning in masked language models from the 2024 \babylmc.
Our results show that even when trained on developmentally plausible quantities of data, models learn diverse constructions, even hard cases that are superficially indistinguishable.
We further find correlational evidence that constructional performance may be functionally relevant: models that better represent constructions perform better on the \babylm benchmarks.\footnote{
\ifrev{All code and data are provided at ANONYMIZED.}
      {All code and data are provided at \href{https://github.com/jsrozner/cxs_are_revealed}{https://github.com/jsrozner/cxs\_are\_revealed}.}
}
\end{abstract}

\section{Introduction}
\label{intro}
Construction Grammars (CxG, \citealt{goldberg1995constructions, goldberg2003constructions, goldberg2006constructions, fillmore1988mechanisms, croft2001radical}) define constructions as form-meaning pairings and typically assume few innate constraints on the inventory of constructions (constructicon).
Thus, a central question in CxG concerns how learners might abstract constructions over time from experience with language (distributional learning;
\citealt{goldberg2003constructions,bybee2006usage,tomasello2005constructing, diessel2004acquisition, diessel2019grammar}).
Some studies have demonstrated the feasibility of distributional learning of constructions in simplified settings \citep{casenhiser2005fast,dunn2017}.
Recent advances in statistical modeling of language \citep{zhao2023survey} have produced pretrained language models (PLMs) that directly instantiate (to a good approximation) the probability distribution over strings (and thus, much of linguistic usage), 
and a growing literature has explored the use of PLMs as tools for testing usage-based linguistic theories \citep{piantadosi2023modern, goldberg2024, milliere2024LMs,weissweiler2025}.

\begin{figure}
    \centering
    \includegraphics[width=\linewidth]{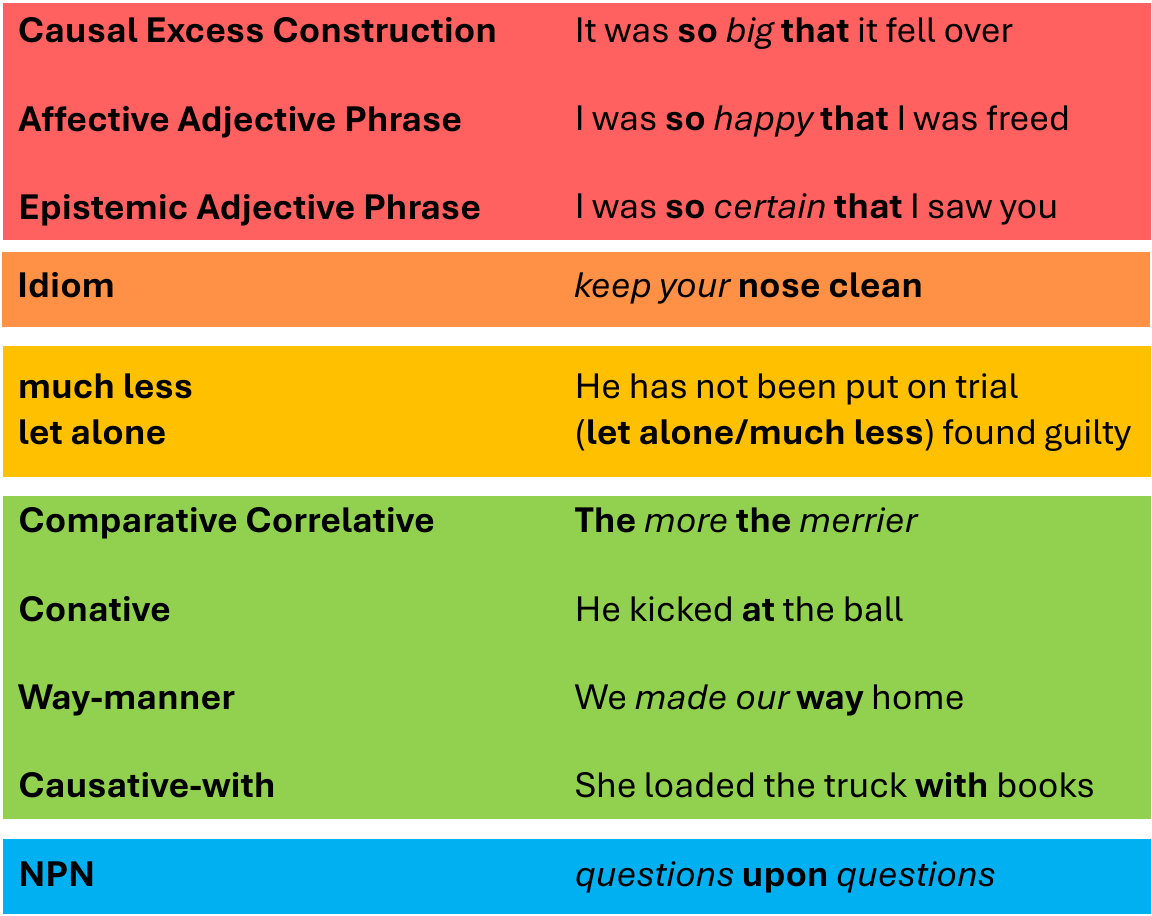}
    \caption{
    Examples of the evaluated constructions.
    \textbf{Bold} reflects \emph{fixed} words and \emph{italics} reflect \emph{schematic} slots constrained to a set of words.
    }
    \label{fig:cxns}
\end{figure}

Recently, motivated by collostructional analysis \citep{stefanowitsch2003} and causal approaches to model study \citep[e.g.,][]{feder-etal-2021-causalm}, 
\citet{rozner2025} hypothesized that constructions should modulate \emph{affinities} (statistical interactions) between words in PLMs' output distributions. 
For example, in the \ttw{let alone} example in \autoref{fig:cxns}, 
\robt assigns 99\% probability to both words when either is masked.
\citeauthor{rozner2025} hypothesized, following related arguments in linguistics 
(\citealt[][p. 248-53]{croft2004cognitive}; \citealt[][p. 169]{hoffmann2022construction}),
that such affinity patterns might hold of constructions more generally. 
They developed causal probing methods and deployed them on \robt \citep{liu2019roberta}, showing that diverse construction types are in fact represented in the distribution \cite[cf.][]{zhou-etal-2024-constructions,bonial-tayyar-madabushi-2024-construction, scivetti2025harish,weissweiler2024hybrid}.

Nonetheless, studies of PLM construction learning have largely used models trained on developmentally implausible quantities of training data.
For example, though the \robt model used by \citeauthor{rozner2025} is less performant than many of the models studied in prior work, it is still trained on roughly 160GB of text ($\sim$30B words),
much more than the 100 million word upper bound estimated for humans by age 13
\cite{hart1997meaningful, gilkerson2017mapping}.
This limits the degree to which patterns of model learning can support inferences about human learning \cite{warstadt2022artificial,frank2023bridging}, and it remains an open question whether models trained on developmentally plausible quantities of data can learn constructions \cite{van-schijndel-etal-2019-quantity, yedetore-etal-2023-poor, Mahowald2024Dissociating, milliere2024LMs}.

In this work, we address the particular question of whether there exist settings---namely choices of architecture, training parameters, and word quantity ($\leq 100M$)---in which statistical learners acquire constructions from developmentally plausible \emph{quantities} of data.
We evaluate eight models from the 2024 \babylmc \citep{hu-etal-2024-findings},
using the experiments from \citeauthor{rozner2025} as a test suite for constructional knowledge.
All evaluations are done in the masked language setting (i.e. with bidirectional context).
Some of the models perform quite well, indicating that developmentally plausible quantities of data are indeed sufficient to recover knowledge of many constructions, as predicted by the usage-based view.
We further find correlational evidence that constructional performance may be functionally relevant: models that better represent constructions perform better on the \babylm evaluations.

\section{Methods}
\label{methods}

\paragraph{Affinity measures}
\citeauthor{rozner2025} developed two \emph{affinity} methods \citep[see also][]{wu-etal-2020-perturbed, hoover-etal-2021-linguistic} and used them to show that constructions manifest as constraints in a \rob's output distribution.
Their affinity measures use bidirectional context (constructions often depend on \emph{subsequent} context), so in this study we test only models that support masked language modeling.

Given a string, $s$,
$s \setminus \mathcal{I}$ is defined as the string with the word indices in $\mathcal{I}$ masked, and 
$\mathcal{P}^i_{s \setminus \mathcal{I}}$ is the \emph{probability distribution} given by the model for the $i$th position in the \emph{masked string}, $s\setminus \mathcal{I}$. 
\emph{Global affinity} is then defined simply as the probability assigned to a word given bidirectional context:
\begin{equation*}
\mathcal{P}^{(i)}_{s\setminus\{i\}}(w_i)
\end{equation*}

\emph{Local affinity} measures pairwise interactions by comparing the change in a model's output distribution for a masked position, 
when another word in the context is also masked,
using Jensen-Shannon Divergence (JSD; \citealt{lin1991divergence}):
\begin{equation*}
a_{i,j} = \text{JSD}(\mathcal{P}^{(j)}_{s\setminus\{j\}}, \mathcal{P}^{(j)}_{s \setminus \{i, j\}})
\end{equation*}

\paragraph{Evaluations}
Evaluations follow \citeauthor{rozner2025} in using affinity to characterize how well the constructions in \autoref{fig:cxns} are reflected in models' output distributions.
Most evaluations look for high global affinities where constructions should constrain word distributions.
Local affinity is used to characterize long-range dependency in the CEC in \sectionrefshort{sec:cec}.
These tests span a wide variety of previously-studied constructions, from fixed (specific-word) to schematic (abstract category), enabling evaluation across degrees of grammatical abstraction:
(i) distinguishing the \emph{causal excess construction} from other constructions with the same surface form, 
(ii) distinguishing literal from figurative usages in potentially idiomatic expressions, 
(iii) recognizing the fixed (substantive) word constraints in six constructions from the Construction Grammar Schematicity corpus (CoGS, \citealt{bonial-tayyar-madabushi-2024-construction}), and 
(iv) recognizing the abstract (schematic) category constraints of the noun-preposition-noun construction (NPN; e.g., \ttw{\textbf{day} after \textbf{day}}, where the bolded slots must be identical nouns) and the comparative correlative (CC).

\paragraph{\babylms used for evaluation}
We test models from the strict and strict-small tracks of the 2024 \babylmc,
in which models were limited to 100M or 10M words, but had no other restrictions on training nor data.
For both tracks, we test the best-performing model, \gptb \citep{charpentier-samuel-2024-bert},
which is a hybrid model that uses both causal and masked language modeling.
We select three additional models from both tracks---roughly the next-best-performing models that also support masked language modeling.
This gives us one \gptb, two \ltgb, and one \rob model for each of the two tracks, so eight total \babylms:
\gptbb and \gptbs \cite{charpentier-samuel-2024-bert}, 
\ltgbb and \ltgbs (the 2024 \babylm baselines; \citealt{hu-etal-2024-findings}), 
\btimeb and \btimes \cite{theodoropoulos-etal-2024-berttime},
and \rob architectures \eli and \qecl \cite{lucas-etal-2024-using, nguyen-etal-2024-automatic}.
We additionally test four non-\babylm models in order to provide an estimate of ceiling performance in our evaluations (subscripts indicate large or base version): \robl (same as tested by \citeauthor{rozner2025}), \robb, \bertl, and \bertb \citep{liu2019roberta, devlin2019bert}.
See \appref{app: babylms} for more details, including \autoref{tab: models} for model descriptions.

\begin{table*}[t]
\relsize{-1}
\begin{NiceTabular}{lm{0.7cm}m{0.5cm}m{0.7cm}*{11}{m{0.5cm}}}
\toprule
\RowStyle{\bfseries}
& & 
\multicolumn{3}{c}{Classification Accuracy} & 
\multicolumn{8}{c}{Global Affinity on Fixed Slots} &
\multicolumn{2}{c}{Schem. Slots} \\

\cmidrule(lr){3-5}
\cmidrule(lr){6-13}
\cmidrule(lr){14-15}

\RowStyle{\bfseries}
& BabyLM & \multicolumn{2}{c}{CEC} & 
\multicolumn{1}{c}{Idioms} &  
\multicolumn{2}{c}{Much Less} &
\multicolumn{2}{c}{Let Alone} &
Con &
Way &
Caus &
CC & 
CC &
NPN  
\\

\cmidrule(lr){2-2}
\cmidrule(lr){3-4}
\cmidrule(lr){5-5}
\cmidrule(lr){6-7}
\cmidrule(lr){8-9}
\cmidrule(lr){10-10}
\cmidrule(lr){11-11}
\cmidrule(lr){12-12}
\cmidrule(lr){13-13}
\cmidrule(lr){14-14}
\cmidrule(lr){15-15}

\RowStyle{\bfseries}
Model & 
\shortstack{Macro\\Avg} & AUC & 
\shortstack{so-\\that} & AUC & 
much & less & let & alone & at &  way  & with & the &
\shortstack{adj/\\adv} &
\shortstack{noun\\ (upon)}
\\ 

\midrule

\gptbb & 75.7 & \colb{94}{93.5} & \colb{94}{93.5} & \colb{55}{55.3} & \colg{62}{62.3} & \colg{52}{52.2} & \colg{95}{94.7} & \colg{94}{94.2} & \colg{19}{19.1} & \colg{63}{62.8} & \colg{92}{91.5} & \colg{92}{91.6} & \colg{100}{99.7} & \colg{81}{81.3} \\ 
\ltgbb & 64.0 & \colb{84}{84.2} & \colb{84}{83.9} & \colb{40}{39.9} & \colg{22}{22.4} & \colg{8}{7.7} & \colg{44}{43.9} & \colg{39}{38.5} & \colg{19}{19.4} & \colg{40}{40.2} & \colg{80}{80.1} & \colg{69}{69.2} & \colg{97}{96.5} & \colg{66}{65.7} \\ 
\btimeb & 63.1 & \colb{85}{85.0} & \colb{87}{87.1} & \colb{34}{34.4} & \colg{10}{10.4} & \colg{5}{5.0} & \colg{2}{1.5} & \colg{4}{3.5} & \colg{14}{14.3} & \colg{34}{34.2} & \colg{87}{87.4} & \colg{93}{92.9} & \colg{94}{93.8} & \colg{43}{42.8} \\ 
\eli & 59.4 & \colb{47}{46.5} & \colb{71}{71.0} & \colb{37}{37.1} & \colg{4}{4.1} & \colg{2}{1.8} & \colg{9}{9.4} & \colg{9}{8.6} & \colg{3}{3.0} & \colg{4}{3.6} & \colg{25}{25.1} & \colg{61}{60.6} & \colg{82}{81.8} & \colg{0}{0.2} \\ 
\midrule
\gptbs & 70.4 & \colb{85}{85.2} & \colb{87}{87.1} & \colb{42}{41.7} & \colg{6}{5.9} & \colg{3}{2.6} & \colg{45}{45.3} & \colg{53}{53.2} & \colg{11}{10.7} & \colg{32}{32.1} & \colg{84}{84.4} & \colg{86}{85.8} & \colg{96}{96.4} & \colg{18}{17.7} \\ 
\btimes & 61.0 & \colb{75}{74.8} & \colb{84}{83.9} & \colb{35}{34.5} & \colg{5}{4.5} & \colg{3}{3.0} & \colg{0}{0.1} & \colg{0}{0.1} & \colg{12}{11.7} & \colg{28}{27.5} & \colg{77}{76.6} & \colg{84}{84.4} & \colg{77}{77.1} & \colg{30}{29.7} \\ 
\qecl & 58.8 & \colb{61}{60.6} & \colb{81}{80.6} & \colb{41}{40.5} & \colg{4}{4.2} & \colg{2}{1.9} & \colg{2}{2.2} & \colg{2}{1.5} & \colg{15}{14.6} & \colg{26}{25.7} & \colg{59}{59.2} & \colg{57}{57.4} & \colg{49}{49.1} & \colg{1}{0.5} \\ 
\ltgbs & 57.6 & \colb{41}{41.1} & \colb{68}{67.7} & \colb{37}{36.6} & \colg{6}{5.6} & \colg{3}{2.8} & \colg{0}{0.2} & \colg{0}{0.2} & \colg{5}{5.2} & \colg{24}{24.2} & \colg{29}{29.1} & \colg{31}{30.9} & \colg{31}{30.7} & \colg{0}{0.3} \\ 
\midrule
\robl &   & \colb{99}{99.4} & \colb{100}{100.0} & \colb{69}{69.2} & \colg{94}{93.5} & \colg{99}{99.1} & \colg{99}{98.5} & \colg{100}{99.6} & \colg{43}{43.4} & \colg{85}{84.8} & \colg{95}{95.0} & \colg{99}{98.9} & \colg{100}{99.9} & \colg{94}{94.4} \\ 
\robb &   & \colb{99}{98.8} & \colb{94}{93.5} & \colb{67}{66.6} & \colg{93}{92.6} & \colg{96}{95.7} & \colg{99}{99.4} & \colg{100}{99.9} & \colg{37}{37.3} & \colg{75}{74.7} & \colg{94}{94.3} & \colg{97}{97.1} & \colg{100}{99.5} & \colg{92}{91.5} \\ 
\bertl &   & \colb{98}{97.9} & \colb{97}{96.8} & \colb{63}{62.5} & \colg{95}{94.8} & \colg{87}{87.1} & \colg{91}{91.3} & \colg{99}{99.3} & \colg{57}{57.3} & \colg{74}{74.1} & \colg{94}{94.1} & \colg{98}{98.0} & \colg{100}{99.9} & \colg{95}{94.6} \\ 
\bertb &   & \colb{96}{96.0} & \colb{100}{100.0} & \colb{57}{56.8} & \colg{79}{79.1} & \colg{77}{76.7} & \colg{88}{88.4} & \colg{95}{95.3} & \colg{50}{49.8} & \colg{66}{65.8} & \colg{90}{89.6} & \colg{99}{98.8} & \colg{98}{98.2} & \colg{86}{85.8} \\ 

\bottomrule

\end{NiceTabular}
\caption{
Results. 
All scores between 0 and 100. 
Blue scores are classifier accuracy; green shows average affinities, except for CC adj/adv.
Left to right: 
BabyLM Macro-Avg reported on HuggingFace;
AUC for classifying CEC vs. EAP/AAP using global aff on \ttw{so}; 
so-that: \% of multithat sentences where \ttw{so} has higher local aff with the causal \ttw{that} than any other \ttw{that};
Idioms: AUC for classifying potentially idiomatic expressions as figurative vs. literal using global aff;
Fixed Slots: Avg global aff for indicated fixed slot;
Schem-CC: For the adj/adv slot, the \% of probability mass that the model places on comparative adj/adv;
Schem-NPN: Avg global aff for nouns in NPNs with P=\ttw{upon}
}
\label{tab: resuls main}
\end{table*}

\section{Experiments and Results}
Results for all experiments are given in \autoref{tab: resuls main}.
Examples of the constructions are found in \autoref{fig:cxns}, and additional experiment details are in \appref{app: eval}.

\subsection{Superficially indistiguishable constructions: the CEC vs. EAP/ AAP} 
\label{sec:cec}
The causal excess construction (CEC; \ttw{I was so happy that I cried}) is superficially indistinguishable from
epistemic and affective adjective phrases (EAP, AAP; \ttw{I was so certain/happy that I saw you}) but admits only \ttw{so} as the adverb: 
\ttw{``I was *very happy that I cried''} has an entirely different meaning \cite{KaySag2012}.
Therefore, models that have learned the CEC should assign high affinity to \ttw{so} in the CEC but not in the EAP/ AAP.
Prior work argued for the difficulty of distinguishing the CEC from the EAP/AAP \citep[][]{zhou-etal-2024-constructions}, but \citeauthor{rozner2025} 
show that the CEC is well-distinguished in the distribution: 
simply thresholding global affinity scores from \robl at 0.78 (i.e., no classifier is trained) correctly characterizes 98\% of examples 
in the \citeauthor{zhou-etal-2024-constructions} CEC dataset.

To quantify whether this distinction is also learned by \babylms, we use the same dataset and compute a receiver operating characteristic (ROC) curve for CEC vs. EAP/AAP classification using global affinity on \ttw{so} as the classifier score (again, the contextual probability of \ttw{so} is directly treated as the classification score).
We report \textbf{CEC AUC}, the area under curve, which reflects how likely models are to assign higher affinity to \ttw{so} in the CEC than in the EAP/AAP.
We also report a \textbf{CEC so-that} score, the percentage of multi-that examples (e.g., \ttw{I was so happy that$_1$ I won \textbf{that$_2$} I smiled})
where \ttw{so} has higher \emph{local affinity} for the causal \ttw{that$_2$} than any distractor \ttw{that}. 
A higher score reflects that a model's distribution for \ttw{so} exhibits the correct (potentially long-range) dependency. 

\paragraph{Results}
The CEC
is well-learned by \gptbb especially, but also by several other models. 
Moreover, so-that accuracy tends to be relatively high even on models that poorly classify the CEC (e.g., \eli and \ltgbs have AUC $< 0.5$ and are thus worse than random), suggesting that models may learn to attend to the correct \ttw{that} (i.e. learn what long range dependencies to attend to) before learning to put most probability mass on \ttw{so} (i.e. become ``confident'' that the CEC requires \ttw{so}).

\subsection{Figurative vs. literal usages}
\citeauthor{rozner2025} show that global affinity provides signal in discriminating literal and figurative usages in potentially idiomatic expressions, 
possibly because the non-compositionality of figurative use leads to greater constraint and thus higher affinity
(e.g., one can \ttw{spill the beans} but in the same context, one would not \ttw{spill the water}, so \ttw{beans} should have high affinity).
Following their approach, we compute \textbf{Idioms AUC} using global affinity to classify figurative vs. literal usages of potentially idiomatic expressions in MAGPIE, a corpus of \mbox{$\sim$50,000} sentences  \citep{haagsma-etal-2020-magpie}.
Higher scores reflect that models on average recognize greater constraint in figurative usages. 
In humans, the long tail of idioms is acquired after other constructions \cite{sprenger2019development, carrol2023old}; here we are interested in corresponding model behavior. 

\paragraph{Results}
Whereas \gptbb achieves fairly good performance on many of our constructional evaluations (\autoref{tab: resuls main}), performance on the MAGPIE dataset is barely above chance (AUC 55.3). 
Perhaps more interesting is that many BabyLMs are substantially \emph{below} chance: 
the worst model, \btimeb, provides a classification signal nearly as good as \robb if the classifier is flipped.
Why might this be the case? 
Though humans (and to some extent \rob) recognize noncompositional uses and their triggering contexts, our results suggest that \babylms trained on less data have not yet learned the long tail of idioms.
It makes sense that uncommon, non-compositional usages would be \emph{surprising} to a \babylm trained on less data and that they would thus exhibit low affinities.

\newcommand{\cxcon}{\textbf{Con}\xspace}
\newcommand{\cxway}{\textbf{Way}\xspace}
\newcommand{\cxcc}{\textbf{CC}\xspace}
\newcommand{\cxcaus}{\textbf{Caus}\xspace}
\newcommand{\cxla}{\textbf{let-alone}\xspace}
\newcommand{\cxml}{\textbf{much-less}\xspace}

\subsection{Fixed slots in partially substantive constructions}
\citeauthor{rozner2025} compute global affinity on the fixed words in $\sim$50 examples of each of six partially schematic constructions (see \autoref{fig:cxns}) from the Construction Grammar Schematicity corpus \citep[CoGs;][]{bonial-tayyar-madabushi-2024-construction} and show that the fixed words often have high global affinities, since the constructional context constrains them.
We evaluate whether \babylms learn to assign high affinities to fixed words in these constructions and whether there are any differences in degree of acquisition.

\paragraph{Results}
\rob converges to nearly 100\% affinity on all fixed words except the \textbf{conative} and \textbf{way-manner}; \citeauthor{rozner2025} note that the conative is relatively rare and that way-manner allows other completions.
\babylms place nontrivial probability on both \ttw{at} and \ttw{way} but still less than the \rob and \bbbb models, suggesting that the idiomatic form of the construction may require greater exposure to become entrenched.
\cxcc and \textbf{causative-with}  are acquired by a number of models, whereas \cxla and \cxml appear harder to learn.
We look more closely at this latter divergence in \sectionrefshort{sec:ml}.

\subsection{Category constraint in the comparative correlative}
Using comparative correlative examples from CoGS we compute the percentage of each model's top-p (p=0.85) completions for the \textbf{CC adj/adv} slot that are comparative (e.g., \ttw{The \textbf{more} the \textbf{merrier}}).
This score represents global affinity at a \emph{category} level, rather than for a single word.
The high scores on numerous models indicate that the abstract category constraint that the slot be an adjective or adverb is well-learned.
In some cases, the abstract constraint seems to be better captured than the fixed word constraint on \ttw{the} (\textbf{CC the}).

\subsection{Generalization of the form of the NPN}
The noun-preposition-noun construction (NPN) is a schematic construction (slots are not constrained to be fixed words, but rather to categories of word---namely two matching nouns and a preposition), so a PLM that shows high affinity for the noun slot has generalized an abstract constraint.
Following \citeauthor{rozner2025} we aim to test generalization to \emph{unseen} NPNs by generating a new dataset of 400 NPN sentences (100 nouns, singly-tokenized in all \babylms for fair comparison, using each of 4 prepositions: after, upon, by, to), 
and the last author, blind to affinity scores, rates the acceptability of each sentence.
We report average global affinity for nouns in the NPN with \ttw{upon} as the preposition (\textbf{NPN noun, upon})
where acceptability is $\geq4$ and where the particular NPN is seen $0$ times in the \gptbb training corpus
(see \appref{app:npn} for additional details and results).
High affinities reflect that models learn that each noun slot must match the other noun; average affinity on the noun slot of 81\% for \emph{unseen} NPNs in \gptbb is a strong signal of acquisition.

\subsection{Much less and let alone: A brief corpus analysis}
\label{sec:ml}
\autoref{tab: resuls main} shows that both \cxla and \cxml are hard for most of the \babylms except for \gptb, which much better learns \cxla than \cxml.
This is interesting because these constructions have similar semantics.
Suprisingly, a review of the \gptbb training data shows that the bigram \ttw{much less} occurs almost twice as often as \ttw{let alone} (765 vs. 439).
However, closer examination reveals that the bigram \ttw{much less} occurs in other settings (e.g., \ttw{John worked \textbf{much less} than Mary}).
We use \rob (which reliably distinguishes \emph{constructional} usages of both \cxla and \cxml; see \autoref{tab: resuls main}) to classify usage (global affinity $\geq$ 0.9 on both words).  
Whereas almost all of the \ttw{let alone} usages are constructional; only 100 ($\sim$13\%) of the \ttw{much less} usages are.
Thus, constructional usage of \emph{much less} is both less frequent and more confusable with other usages, likely impeding acquisition relative to \emph{let alone}.

\subsection{Better construction learning is associated with better downstream performance}
To assess the functional relevance of construction learning, we include the \textbf{\babylm macro average} in the first column (see \autoref{tab: app: babylm score} for details of this score).
We compute the correlation of each construction score with the \babylm macro average and then average, giving an average correlation of $r=0.78\pm0.10$ SD,
which shows that performance on the constructional tests in the masked language setting correlates with \babylm performance.

\section{Related Work and Discussion}
Prior work on grammatical knowledge in cognitively plausible models has tended to focus on syntactic minimal pairs \cite{warstadt-etal-2023-findings, wilcox2025bigger, bunzeck-etal-2025-construction, marvin-linzen-2018-targeted} using datasets like BLiMP and CLAMS \cite{warstadt-etal-2020-blimp-benchmark, mueller-etal-2020-cross}.
Studies of more complex constructions have tended to use probing and prompting (\citealt{rozner2025}; see also \citealp{weissweiler-etal-2023-construction, milliere2024LMs} for discussion).
More recently, \babylm-like models have been trained on curated corpora to test how relatively rare constructions are acquired from limited data \citep[e.g.,][]{misra-mahowald-2024-language, leong2024}.
Though prior work has questioned whether complex constructions can be learned from cognitively plausible quantities of data \cite{van-schijndel-etal-2019-quantity, yedetore-etal-2023-poor, Mahowald2024Dissociating, milliere2024LMs},
our results provide evidence that even difficult constructions can be acquired from developmentally plausible quantities of data and that the extent of acquisition correlates with general performance.

Prior work has considered the patterns of acquisition of syntactic capacities over model training \cite{zhang-etal-2021-need, warstadt2022artificial,wilcox2025bigger}.
In this work, we see that when compared to \robt, a larger model trained on $300-3000\times$ more data, \babylms show divergences in their acquisition of different constructions.
In more recent work \citet{bunzeck-etal-2025-construction} report that the distribution of constructions (basic sentence types; e.g., wh-question, copula, imperative) does not substantially impact learning trajectories when models are evaluated on lexical, syntactic, and semantic minimal pairs; although
\citet{bunzeck-zarriess-2024-fifty} provide evidence that the shape of learning curves does vary across different syntactic phenomena in BLiMP.
In this study we do find a correlation between corpus composition and performance on \ttw{much less} and \ttw{let alone}, which makes sense, given that substantive constructions with fixed words must be seen to be learned.
The low AUC we observe in \babylms for classifying potentially idiomatic expressions likely reflects the same effect: learning the long tail of idioms requires exposure.
Future work should investigate the interplay between corpus composition and acquisition dynamics across the spectrum of constructions.

\section{Conclusion}
Models trained on cognitively plausible quantities of data acquire diverse constructions.
This result provides empirical support for the feasibility of distributional learning.
Moreover, we found that differences in construction acquisition in \babylms exhibit some similarities to human learning.
Prior studies of \babylms have tended to focus on simpler constructional distinctions via minimal pairs; 
here we used targeted distributional evaluation to study acquisition of more complex constructions. 
Given that acquisition correlates with other functional behaviors, 
future work should examine acquisition dynamics, including interactions between simple and complex constructions during learning.

\section*{Limitations}
This is a computational modeling study and thus comes with the usual caveats: the ``subjects'' are models, and inferences to humans should be drawn with care.
This study improves on the cognitive plausibility of prior work only in the \emph{amount}---but not the \emph{kind}---of linguistic experience;
some recent findings suggest that content may play surprisingly little role in shaping LMs’ linguistic abstractions \cite{feng-etal-2024-child}.
100M words corresponds roughly to the number of words seen by a 13 year-old,
and we consider it likely that 13 year-olds have acquired many of the constructions we study in this paper, except possibly rare idioms,
though further work might compare the course of acquisition in humans and \babylms.

We evaluated only models that support bidirectional context, which is  implausible for process models of language comprehension  \citep[e.g.,][]{frazier1978sausage}, as the constructions we evaluate depend on subsequent context.
Models also differ from humans in other important ways including learning dynamics and architecture \citep[see e.g.,][]{frank2023bridging}.
Though likelihood scores over entire sentences do correlate with human grammaticality judgements \cite{hu2024language}, no direct work has yet been done to correlate bidirectional affinity scores with human behaviors.

Our study focused on whether model distributions reflect certain formal and semantic distinctions between constructions, but does not allow us to conclude that models ``know'' constructions in every sense relevant to humans (e.g., that they can recognize and reason about their truth conditions; see e.g., \citealt{zhou-etal-2024-constructions, weissweiler-etal-2022-better} for countervailing evidence). 
While we leave study of these dimensions of semantic knowledge to future work, we believe that the ability to make the distinctions we study is a critical component of grammar learning and highly relevant to the distributional hypothesis for human language acquisition \citep[see e.g.,][p. 30]{tomasello2005constructing}.

Finally, we have framed our study against the theoretical background of construction grammar because we are motivated by considerations about learning that arise from that background.
Our results do not in themselves argue for CxG over other theories of natural language syntax, and we do not mean to imply that our findings cannot be accommodated by other views of the nature of human linguistic knowledge.
We have simply provided evidence that much constructional information is available to a sufficiently performant statistical learner, as is typically required by CxG theories.

\section*{Acknowledgements}
Leonie Weissweiler was supported by a postdoctoral fellowship from the German Research Foundation (DFG, \texttt{WE 7627/1-1}). 

\bibliography{josh, anthology1, anthology2, literatur_25_03} 

\begin{thebibliography}{65}
\providecommand{\natexlab}[1]{#1}

\bibitem[{Bonial and Tayyar~Madabushi(2024)}]{bonial-tayyar-madabushi-2024-construction}
Claire Bonial and Harish Tayyar~Madabushi. 2024.
\newblock \href {https://aclanthology.org/2024.lrec-main.22/} {A construction grammar corpus of varying schematicity: A dataset for the evaluation of abstractions in language models}.
\newblock In \emph{Proceedings of the 2024 Joint International Conference on Computational Linguistics, Language Resources and Evaluation (LREC-COLING 2024)}, pages 243--255, Torino, Italia. ELRA and ICCL.

\bibitem[{Bunzeck et~al.(2025)Bunzeck, Duran, and Zarrie{\ss}}]{bunzeck-etal-2025-construction}
Bastian Bunzeck, Daniel Duran, and Sina Zarrie{\ss}. 2025.
\newblock \href {https://doi.org/10.18653/v1/2025.conll-1.12} {Do construction distributions shape formal language learning in {G}erman {B}aby{LM}s?}
\newblock In \emph{Proceedings of the 29th Conference on Computational Natural Language Learning}, pages 169--186, Vienna, Austria. Association for Computational Linguistics.

\bibitem[{Bunzeck and Zarrie{\ss}(2024)}]{bunzeck-zarriess-2024-fifty}
Bastian Bunzeck and Sina Zarrie{\ss}. 2024.
\newblock \href {https://aclanthology.org/2024.clasp-1.7/} {Fifty shapes of {BL}i{MP}: syntactic learning curves in language models are not uniform, but sometimes unruly}.
\newblock In \emph{Proceedings of the 2024 CLASP Conference on Multimodality and Interaction in Language Learning}, pages 39--55, Gothenburg, Sweden. Association for Computational Linguistics.

\bibitem[{Bybee(2006)}]{bybee2006usage}
Joan Bybee. 2006.
\newblock From usage to grammar: The mind's response to repetition.
\newblock \emph{Language}, pages 711--733.

\bibitem[{Carrol(2023)}]{carrol2023old}
Gareth Carrol. 2023.
\newblock Old dogs and new tricks: Assessing idiom knowledge amongst native speakers of different ages.
\newblock \emph{Journal of Psycholinguistic Research}, 52(6):2287--2302.

\bibitem[{Casenhiser and Goldberg(2005)}]{casenhiser2005fast}
Devin Casenhiser and Adele~E Goldberg. 2005.
\newblock Fast mapping between a phrasal form and meaning.
\newblock \emph{Developmental science}, 8(6):500--508.

\bibitem[{Charpentier et~al.(2025)Charpentier, Choshen, Cotterell, Gul, Hu, Jumelet, Linzen, Liu, Mueller, Ross, Shah, Warstadt, Wilcox, and Williams}]{charpentier2025babylm}
Lucas Charpentier, Leshem Choshen, Ryan Cotterell, Mustafa~Omer Gul, Michael Hu, Jaap Jumelet, Tal Linzen, Jing Liu, Aaron Mueller, Candace Ross, Raj~Sanjay Shah, Alex Warstadt, Ethan Wilcox, and Adina Williams. 2025.
\newblock \href {https://arxiv.org/abs/2502.10645} {Babylm turns 3: Call for papers for the 2025 babylm workshop}.
\newblock \emph{Preprint}, arXiv:2502.10645.

\bibitem[{Charpentier and Samuel(2024)}]{charpentier-samuel-2024-bert}
Lucas Georges~Gabriel Charpentier and David Samuel. 2024.
\newblock \href {https://aclanthology.org/2024.conll-babylm.24/} {{GPT} or {BERT}: why not both?}
\newblock In \emph{The 2nd BabyLM Challenge at the 28th Conference on Computational Natural Language Learning}, pages 262--283, Miami, FL, USA. Association for Computational Linguistics.

\bibitem[{Croft(2001)}]{croft2001radical}
William Croft. 2001.
\newblock \emph{Radical construction grammar: Syntactic theory in typological perspective}.
\newblock Oxford University Press, USA.

\bibitem[{Croft and Cruse(2004)}]{croft2004cognitive}
William Croft and D~Alan Cruse. 2004.
\newblock \emph{Cognitive linguistics}.
\newblock Cambridge University Press.

\bibitem[{Devlin et~al.(2019)Devlin, Chang, Lee, and Toutanova}]{devlin2019bert}
Jacob Devlin, Ming-Wei Chang, Kenton Lee, and Kristina Toutanova. 2019.
\newblock \href {https://arxiv.org/abs/1810.04805} {Bert: Pre-training of deep bidirectional transformers for language understanding}.
\newblock \emph{Preprint}, arXiv:1810.04805.

\bibitem[{Diessel(2004)}]{diessel2004acquisition}
Holger Diessel. 2004.
\newblock \emph{The acquisition of complex sentences}.
\newblock Cambridge University Press.

\bibitem[{Diessel(2019)}]{diessel2019grammar}
Holger Diessel. 2019.
\newblock \emph{The grammar network}.
\newblock Cambridge University Press.

\bibitem[{Dunn(2017)}]{dunn2017}
Jonathan Dunn. 2017.
\newblock Computational learning of construction grammars.
\newblock \emph{Language and cognition}, 9(2):254--292.

\bibitem[{Feder et~al.(2021)Feder, Oved, Shalit, and Reichart}]{feder-etal-2021-causalm}
Amir Feder, Nadav Oved, Uri Shalit, and Roi Reichart. 2021.
\newblock \href {https://doi.org/10.1162/coli_a_00404} {{C}ausa{LM}: Causal model explanation through counterfactual language models}.
\newblock \emph{Computational Linguistics}, 47(2):333--386.

\bibitem[{Feng et~al.(2024)Feng, Goodman, and Frank}]{feng-etal-2024-child}
Steven~Y. Feng, Noah~D. Goodman, and Michael~C. Frank. 2024.
\newblock \href {https://doi.org/10.18653/v1/2024.emnlp-main.1231} {Is child-directed speech effective training data for language models?}
\newblock In \emph{Proceedings of the 2024 Conference on Empirical Methods in Natural Language Processing}, pages 22055--22071, Miami, Florida, USA. Association for Computational Linguistics.

\bibitem[{Fillmore(1988)}]{fillmore1988mechanisms}
Charles~J Fillmore. 1988.
\newblock The mechanisms of ``construction grammar".
\newblock In \emph{Annual Meeting of the Berkeley Linguistics Society}, volume~14, pages 35--55.

\bibitem[{Frank(2023)}]{frank2023bridging}
Michael~C Frank. 2023.
\newblock Bridging the data gap between children and large language models.
\newblock \emph{Trends in Cognitive Sciences}.

\bibitem[{Frazier and Fodor(1978)}]{frazier1978sausage}
Lyn Frazier and Janet~Dean Fodor. 1978.
\newblock The sausage machine: A new two-stage parsing model.
\newblock \emph{Cognition}, 6(4):291--325.

\bibitem[{Gilkerson et~al.(2017)Gilkerson, Richards, Warren, Montgomery, Greenwood, Kimbrough~Oller, Hansen, and Paul}]{gilkerson2017mapping}
Jill Gilkerson, Jeffrey~A Richards, Steven~F Warren, Judith~K Montgomery, Charles~R Greenwood, D~Kimbrough~Oller, John~HL Hansen, and Terrance~D Paul. 2017.
\newblock Mapping the early language environment using all-day recordings and automated analysis.
\newblock \emph{American journal of speech-language pathology}, 26(2):248--265.

\bibitem[{Goldberg(1995)}]{goldberg1995constructions}
Adele~E Goldberg. 1995.
\newblock \emph{Constructions: A construction grammar approach to argument structure}.
\newblock University of Chicago Press.

\bibitem[{Goldberg(2003)}]{goldberg2003constructions}
Adele~E Goldberg. 2003.
\newblock Constructions: A new theoretical approach to language.
\newblock \emph{Trends in cognitive sciences}, 7(5):219--224.

\bibitem[{Goldberg(2006)}]{goldberg2006constructions}
Adele~E Goldberg. 2006.
\newblock \emph{Constructions at Work: The Nature of Generalization in Language}.
\newblock Oxford University Press, USA.

\bibitem[{Goldberg(2024)}]{goldberg2024}
Adele~E. Goldberg. 2024.
\newblock \href {https://doi.org/10.1075/cf.23017.gol} {Usage-based constructionist approaches and large language models}.
\newblock \emph{Constructions and Frames}, 16(2):220--254.

\bibitem[{Haagsma et~al.(2020)Haagsma, Bos, and Nissim}]{haagsma-etal-2020-magpie}
Hessel Haagsma, Johan Bos, and Malvina Nissim. 2020.
\newblock \href {https://aclanthology.org/2020.lrec-1.35/} {{MAGPIE}: A large corpus of potentially idiomatic expressions}.
\newblock In \emph{Proceedings of the Twelfth Language Resources and Evaluation Conference}, pages 279--287, Marseille, France. European Language Resources Association.

\bibitem[{Hart et~al.(1997)Hart, Risley, and Kirby}]{hart1997meaningful}
Betty Hart, Todd~R Risley, and John~R Kirby. 1997.
\newblock Meaningful differences in the everyday experience of young american children.
\newblock \emph{Canadian Journal of Education}, 22(3):323.

\bibitem[{Hoffmann(2022)}]{hoffmann2022construction}
Thomas Hoffmann. 2022.
\newblock \emph{Construction grammar}.
\newblock Cambridge University Press.

\bibitem[{Honnibal et~al.(2020)Honnibal, Montani, Van~Landeghem, and Boyd}]{spacy}
Matthew Honnibal, Ines Montani, Sofie Van~Landeghem, and Adriane Boyd. 2020.
\newblock \href {https://doi.org/10.5281/zenodo.1212303} {{spaCy}: Industrial-strength natural language processing in python}.

\bibitem[{Hoover et~al.(2021)Hoover, Du, Sordoni, and O{'}Donnell}]{hoover-etal-2021-linguistic}
Jacob~Louis Hoover, Wenyu Du, Alessandro Sordoni, and Timothy~J. O{'}Donnell. 2021.
\newblock \href {https://doi.org/10.18653/v1/2021.emnlp-main.234} {Linguistic dependencies and statistical dependence}.
\newblock In \emph{Proceedings of the 2021 Conference on Empirical Methods in Natural Language Processing}, pages 2941--2963, Online and Punta Cana, Dominican Republic. Association for Computational Linguistics.

\bibitem[{Hu et~al.(2024{\natexlab{a}})Hu, Mahowald, Lupyan, Ivanova, and Levy}]{hu2024language}
Jennifer Hu, Kyle Mahowald, Gary Lupyan, Anna Ivanova, and Roger Levy. 2024{\natexlab{a}}.
\newblock \href {https://doi.org/10.1073/pnas.2400917121} {Language {M}odels {A}lign with {H}uman {J}udgments on {K}ey {G}rammatical {C}onstructions}.
\newblock \emph{Proceedings of the National Academy of Sciences}, 121(36):e2400917121.

\bibitem[{Hu et~al.(2024{\natexlab{b}})Hu, Mueller, Ross, Williams, Linzen, Zhuang, Cotterell, Choshen, Warstadt, and Wilcox}]{hu-etal-2024-findings}
Michael~Y. Hu, Aaron Mueller, Candace Ross, Adina Williams, Tal Linzen, Chengxu Zhuang, Ryan Cotterell, Leshem Choshen, Alex Warstadt, and Ethan~Gotlieb Wilcox. 2024{\natexlab{b}}.
\newblock \href {https://aclanthology.org/2024.conll-babylm.1/} {Findings of the second {B}aby{LM} challenge: Sample-efficient pretraining on developmentally plausible corpora}.
\newblock In \emph{The 2nd BabyLM Challenge at the 28th Conference on Computational Natural Language Learning}, pages 1--21, Miami, FL, USA. Association for Computational Linguistics.

\bibitem[{Kay and Sag(2012)}]{KaySag2012}
Paul Kay and Ivan~A Sag. 2012.
\newblock Cleaning up the big mess: Discontinuous dependencies and complex determiners.
\newblock In \emph{Sign-based construction grammar}, chapter~5, pages 229--256. Citeseer.

\bibitem[{Leong and Linzen(2024)}]{leong2024}
Cara Su-Yi Leong and Tal Linzen. 2024.
\newblock \href {https://arxiv.org/abs/2407.04593} {Testing learning hypotheses using neural networks by manipulating learning data}.
\newblock \emph{Preprint}, arXiv:2407.04593.

\bibitem[{Lin(1991)}]{lin1991divergence}
Jianhua Lin. 1991.
\newblock Divergence measures based on the shannon entropy.
\newblock \emph{IEEE Transactions on Information theory}, 37(1):145--151.

\bibitem[{Liu et~al.(2019)Liu, Ott, Goyal, Du, Joshi, Chen, Levy, Lewis, Zettlemoyer, and Stoyanov}]{liu2019roberta}
Yinhan Liu, Myle Ott, Naman Goyal, Jingfei Du, Mandar Joshi, Danqi Chen, Omer Levy, Mike Lewis, Luke Zettlemoyer, and Veselin Stoyanov. 2019.
\newblock \href {https://arxiv.org/abs/1907.11692} {{RoBERTa}: A robustly optimized bert pretraining approach}.
\newblock \emph{Preprint}, arXiv:1907.11692.

\bibitem[{Lucas et~al.(2024)Lucas, Gaines, Kosireddy, Li, and Havens}]{lucas-etal-2024-using}
Evan Lucas, Dylan Gaines, Tagore~Rao Kosireddy, Kevin Li, and Timothy~C. Havens. 2024.
\newblock \href {https://aclanthology.org/2024.conll-babylm.19/} {Using curriculum masking based on child language development to train a large language model with limited training data}.
\newblock In \emph{The 2nd BabyLM Challenge at the 28th Conference on Computational Natural Language Learning}, pages 221--228, Miami, FL, USA. Association for Computational Linguistics.

\bibitem[{Mahowald et~al.(2024)Mahowald, Ivanova, Blank, Kanwisher, Tenenbaum, and Fedorenko}]{Mahowald2024Dissociating}
Kyle Mahowald, Anna~A Ivanova, Idan~A Blank, Nancy Kanwisher, Joshua~B Tenenbaum, and Evelina Fedorenko. 2024.
\newblock Dissociating language and thought in large language models.
\newblock \emph{Trends in Cognitive Sciences}, 28(6):517--540.

\bibitem[{Marvin and Linzen(2018)}]{marvin-linzen-2018-targeted}
Rebecca Marvin and Tal Linzen. 2018.
\newblock \href {https://doi.org/10.18653/v1/D18-1151} {Targeted syntactic evaluation of language models}.
\newblock In \emph{Proceedings of the 2018 Conference on Empirical Methods in Natural Language Processing}, pages 1192--1202, Brussels, Belgium. Association for Computational Linguistics.

\bibitem[{Millière(2024)}]{milliere2024LMs}
Raphaël Millière. 2024.
\newblock \href {https://arxiv.org/abs/2408.07144} {Language models as models of language}.
\newblock \emph{Preprint}, arXiv:2408.07144.

\bibitem[{Misra and Mahowald(2024)}]{misra-mahowald-2024-language}
Kanishka Misra and Kyle Mahowald. 2024.
\newblock \href {https://doi.org/10.18653/v1/2024.emnlp-main.53} {Language models learn rare phenomena from less rare phenomena: The case of the missing {AANN}s}.
\newblock In \emph{Proceedings of the 2024 Conference on Empirical Methods in Natural Language Processing}, pages 913--929, Miami, Florida, USA. Association for Computational Linguistics.

\bibitem[{Mueller et~al.(2020)Mueller, Nicolai, Petrou-Zeniou, Talmina, and Linzen}]{mueller-etal-2020-cross}
Aaron Mueller, Garrett Nicolai, Panayiota Petrou-Zeniou, Natalia Talmina, and Tal Linzen. 2020.
\newblock \href {https://doi.org/10.18653/v1/2020.acl-main.490} {Cross-linguistic syntactic evaluation of word prediction models}.
\newblock In \emph{Proceedings of the 58th Annual Meeting of the Association for Computational Linguistics}, pages 5523--5539, Online. Association for Computational Linguistics.

\bibitem[{Nguyen et~al.(2024)Nguyen, Yip, and DeBenedetto}]{nguyen-etal-2024-automatic}
Hiep Nguyen, Lynn Yip, and Justin DeBenedetto. 2024.
\newblock \href {https://aclanthology.org/2024.conll-babylm.18/} {Automatic quality estimation for data selection and curriculum learning}.
\newblock In \emph{The 2nd BabyLM Challenge at the 28th Conference on Computational Natural Language Learning}, pages 212--220, Miami, FL, USA. Association for Computational Linguistics.

\bibitem[{Piantadosi(2024)}]{piantadosi2023modern}
Steven~T. Piantadosi. 2024.
\newblock Modern language models refute chomsky’s approach to language.
\newblock In Edward Gibson and Moshe Poliak, editors, \emph{From fieldwork to linguistic theory: A tribute to Dan Everett}. Language Science Press.

\bibitem[{Rozner et~al.(2025)Rozner, Weissweiler, Mahowald, and Shain}]{rozner2025}
Joshua Rozner, Leonie Weissweiler, Kyle Mahowald, and Cory Shain. 2025.
\newblock Constructions are revealed in word distributions.
\newblock In \emph{The 2025 Conference on Empirical Methods in Natural Language Processing}.
\newblock To appear.

\bibitem[{Samuel et~al.(2023)Samuel, Kutuzov, {\O}vrelid, and Velldal}]{samuel-etal-2023-trained}
David Samuel, Andrey Kutuzov, Lilja {\O}vrelid, and Erik Velldal. 2023.
\newblock \href {https://doi.org/10.18653/v1/2023.findings-eacl.146} {Trained on 100 million words and still in shape: {BERT} meets {B}ritish {N}ational {C}orpus}.
\newblock In \emph{Findings of the Association for Computational Linguistics: EACL 2023}, pages 1954--1974, Dubrovnik, Croatia. Association for Computational Linguistics.

\bibitem[{Scivetti et~al.(2025)Scivetti, Torgbi, Blodgett, Shichman, Hudson, Bonial, and Madabushi}]{scivetti2025harish}
Wesley Scivetti, Melissa Torgbi, Austin Blodgett, Mollie Shichman, Taylor Hudson, Claire Bonial, and Harish~Tayyar Madabushi. 2025.
\newblock \href {https://doi.org/10.48550/arXiv.2501.04661} {Assessing language comprehension in large language models using construction grammar}.
\newblock \emph{CoRR}, abs/2501.04661.

\bibitem[{Sprenger et~al.(2019)Sprenger, la~Roi, and Van~Rij}]{sprenger2019development}
Simone~A Sprenger, Am{\'e}lie la~Roi, and Jacolien Van~Rij. 2019.
\newblock The development of idiom knowledge across the lifespan.
\newblock \emph{Frontiers in Communication}, 4:29.

\bibitem[{Stefanowitsch and Gries(2003)}]{stefanowitsch2003}
Anatol Stefanowitsch and Stefan~Th. Gries. 2003.
\newblock \href {https://doi.org/10.1075/ijcl.8.2.03ste} {Collostructions: Investigating the interaction of words and constructions}.
\newblock \emph{International Journal of Corpus Linguistics}, 8(2):209--243.

\bibitem[{Theodoropoulos et~al.(2024)Theodoropoulos, Filandrianos, Lyberatos, Lymperaiou, and Stamou}]{theodoropoulos-etal-2024-berttime}
Nikitas Theodoropoulos, Giorgos Filandrianos, Vassilis Lyberatos, Maria Lymperaiou, and Giorgos Stamou. 2024.
\newblock \href {https://aclanthology.org/2024.conll-babylm.28/} {{BERT}time stories: Investigating the role of synthetic story data in language pre-training}.
\newblock In \emph{The 2nd BabyLM Challenge at the 28th Conference on Computational Natural Language Learning}, pages 308--323, Miami, FL, USA. Association for Computational Linguistics.

\bibitem[{Tomasello(2005)}]{tomasello2005constructing}
Michael Tomasello. 2005.
\newblock \emph{Constructing a language: A usage-based theory of language acquisition}.
\newblock Harvard university press.

\bibitem[{van Schijndel et~al.(2019)van Schijndel, Mueller, and Linzen}]{van-schijndel-etal-2019-quantity}
Marten van Schijndel, Aaron Mueller, and Tal Linzen. 2019.
\newblock \href {https://doi.org/10.18653/v1/D19-1592} {Quantity doesn{'}t buy quality syntax with neural language models}.
\newblock In \emph{Proceedings of the 2019 Conference on Empirical Methods in Natural Language Processing and the 9th International Joint Conference on Natural Language Processing (EMNLP-IJCNLP)}, pages 5831--5837, Hong Kong, China. Association for Computational Linguistics.

\bibitem[{Warstadt and Bowman(2022)}]{warstadt2022artificial}
Alex Warstadt and Samuel~R Bowman. 2022.
\newblock What artificial neural networks can tell us about human language acquisition.
\newblock In \emph{Algebraic structures in natural language}, pages 17--60. CRC Press.

\bibitem[{Warstadt et~al.(2023)Warstadt, Mueller, Choshen, Wilcox, Zhuang, Ciro, Mosquera, Paranjabe, Williams, Linzen, and Cotterell}]{warstadt-etal-2023-findings}
Alex Warstadt, Aaron Mueller, Leshem Choshen, Ethan Wilcox, Chengxu Zhuang, Juan Ciro, Rafael Mosquera, Bhargavi Paranjabe, Adina Williams, Tal Linzen, and Ryan Cotterell. 2023.
\newblock \href {https://doi.org/10.18653/v1/2023.conll-babylm.1} {Findings of the {B}aby{LM} challenge: Sample-efficient pretraining on developmentally plausible corpora}.
\newblock In \emph{Proceedings of the BabyLM Challenge at the 27th Conference on Computational Natural Language Learning}, pages 1--34, Singapore. Association for Computational Linguistics.

\bibitem[{Warstadt et~al.(2020)Warstadt, Parrish, Liu, Mohananey, Peng, Wang, and Bowman}]{warstadt-etal-2020-blimp-benchmark}
Alex Warstadt, Alicia Parrish, Haokun Liu, Anhad Mohananey, Wei Peng, Sheng-Fu Wang, and Samuel~R. Bowman. 2020.
\newblock \href {https://doi.org/10.1162/tacl_a_00321} {{BL}i{MP}: The benchmark of linguistic minimal pairs for {E}nglish}.
\newblock \emph{Transactions of the Association for Computational Linguistics}, 8:377--392.

\bibitem[{Weissweiler et~al.(2023)Weissweiler, He, Otani, R.~Mortensen, Levin, and Sch{\"u}tze}]{weissweiler-etal-2023-construction}
Leonie Weissweiler, Taiqi He, Naoki Otani, David R.~Mortensen, Lori Levin, and Hinrich Sch{\"u}tze. 2023.
\newblock \href {https://aclanthology.org/2023.cxgsnlp-1.10/} {Construction grammar provides unique insight into neural language models}.
\newblock In \emph{Proceedings of the First International Workshop on Construction Grammars and NLP (CxGs+NLP, GURT/SyntaxFest 2023)}, pages 85--95, Washington, D.C. Association for Computational Linguistics.

\bibitem[{Weissweiler et~al.(2022)Weissweiler, Hofmann, K{\"o}ksal, and Sch{\"u}tze}]{weissweiler-etal-2022-better}
Leonie Weissweiler, Valentin Hofmann, Abdullatif K{\"o}ksal, and Hinrich Sch{\"u}tze. 2022.
\newblock \href {https://doi.org/10.18653/v1/2022.emnlp-main.746} {The better your syntax, the better your semantics? probing pretrained language models for the {E}nglish comparative correlative}.
\newblock In \emph{Proceedings of the 2022 Conference on Empirical Methods in Natural Language Processing}, pages 10859--10882, Abu Dhabi, United Arab Emirates. Association for Computational Linguistics.

\bibitem[{Weissweiler et~al.(2024)Weissweiler, Köksal, and Schütze}]{weissweiler2024hybrid}
Leonie Weissweiler, Abdullatif Köksal, and Hinrich Schütze. 2024.
\newblock \href {https://arxiv.org/abs/2403.06965} {Hybrid human-{LLM} corpus construction and {LLM} evaluation for rare linguistic phenomena}.
\newblock \emph{Preprint}, arXiv:2403.06965.

\bibitem[{Weissweiler et~al.(2025)Weissweiler, Mahowald, and Goldberg}]{weissweiler2025}
Leonie Weissweiler, Kyle Mahowald, and Adele Goldberg. 2025.
\newblock \href {https://arxiv.org/abs/2502.13195} {Linguistic generalizations are not rules: Impacts on evaluation of lms}.
\newblock \emph{Preprint}, arXiv:2502.13195.

\bibitem[{Wilcox et~al.(2025)Wilcox, Hu, Mueller, Linzen, Warstadt, Choshen, Zhuang, Cotterell, and Williams}]{wilcox2025bigger}
Ethan~G Wilcox, Michael Hu, Aaron Mueller, Tal Linzen, Alex Warstadt, Leshem Choshen, Chengxu Zhuang, Ryan Cotterell, and Adina Williams. 2025.
\newblock \href {https://doi.org/10.31234/osf.io/rfwgd_v2} {Bigger is not always better: The importance of human-scale language modeling for psycholinguistics}.

\bibitem[{Wolf et~al.(2019)Wolf, Debut, Sanh, Chaumond, Delangue, Moi, Cistac, Rault, Louf, Funtowicz, and Brew}]{huggingface}
Thomas Wolf, Lysandre Debut, Victor Sanh, Julien Chaumond, Clement Delangue, Anthony Moi, Pierric Cistac, Tim Rault, R{\'{e}}mi Louf, Morgan Funtowicz, and Jamie Brew. 2019.
\newblock \href {https://arxiv.org/abs/1910.03771} {Huggingface's transformers: State-of-the-art natural language processing}.
\newblock \emph{CoRR}, abs/1910.03771.

\bibitem[{Wu et~al.(2020)Wu, Chen, Kao, and Liu}]{wu-etal-2020-perturbed}
Zhiyong Wu, Yun Chen, Ben Kao, and Qun Liu. 2020.
\newblock \href {https://doi.org/10.18653/v1/2020.acl-main.383} {Perturbed masking: Parameter-free probing for analyzing and interpreting {BERT}}.
\newblock In \emph{Proceedings of the 58th Annual Meeting of the Association for Computational Linguistics}, pages 4166--4176, Online. Association for Computational Linguistics.

\bibitem[{Yedetore et~al.(2023)Yedetore, Linzen, Frank, and McCoy}]{yedetore-etal-2023-poor}
Aditya Yedetore, Tal Linzen, Robert Frank, and R.~Thomas McCoy. 2023.
\newblock \href {https://doi.org/10.18653/v1/2023.acl-long.521} {How poor is the stimulus? evaluating hierarchical generalization in neural networks trained on child-directed speech}.
\newblock In \emph{Proceedings of the 61st Annual Meeting of the Association for Computational Linguistics (Volume 1: Long Papers)}, pages 9370--9393, Toronto, Canada. Association for Computational Linguistics.

\bibitem[{Zhang et~al.(2021)Zhang, Warstadt, Li, and Bowman}]{zhang-etal-2021-need}
Yian Zhang, Alex Warstadt, Xiaocheng Li, and Samuel~R. Bowman. 2021.
\newblock \href {https://doi.org/10.18653/v1/2021.acl-long.90} {When do you need billions of words of pretraining data?}
\newblock In \emph{Proceedings of the 59th Annual Meeting of the Association for Computational Linguistics and the 11th International Joint Conference on Natural Language Processing (Volume 1: Long Papers)}, pages 1112--1125, Online. Association for Computational Linguistics.

\bibitem[{Zhao et~al.(2025)Zhao, Zhou, Li, Tang, Wang, Hou, Min, Zhang, Zhang, Dong, Du, Yang, Chen, Chen, Jiang, Ren, Li, Tang, Liu, Liu, Nie, and Wen}]{zhao2023survey}
Wayne~Xin Zhao, Kun Zhou, Junyi Li, Tianyi Tang, Xiaolei Wang, Yupeng Hou, Yingqian Min, Beichen Zhang, Junjie Zhang, Zican Dong, Yifan Du, Chen Yang, Yushuo Chen, Zhipeng Chen, Jinhao Jiang, Ruiyang Ren, Yifan Li, Xinyu Tang, Zikang Liu, and 3 others. 2025.
\newblock \href {https://arxiv.org/abs/2303.18223} {A survey of large language models}.
\newblock \emph{Preprint}, arXiv:2303.18223.

\bibitem[{Zhou et~al.(2024)Zhou, Weissweiler, He, Sch{\"u}tze, Mortensen, and Levin}]{zhou-etal-2024-constructions}
Shijia Zhou, Leonie Weissweiler, Taiqi He, Hinrich Sch{\"u}tze, David~R. Mortensen, and Lori Levin. 2024.
\newblock \href {https://aclanthology.org/2024.lrec-main.336/} {Constructions are so difficult that {E}ven large language models get them right for the wrong reasons}.
\newblock In \emph{Proceedings of the 2024 Joint International Conference on Computational Linguistics, Language Resources and Evaluation (LREC-COLING 2024)}, pages 3804--3811, Torino, Italia. ELRA and ICCL.

\end{thebibliography}

\newpage
\appendix

\begin{table*}[t]
\begin{NiceTabular}{llrrlp{4cm}}
\toprule

\RowStyle{\bfseries}
& & & 
\multicolumn{2}{c}{Training Data}  & \\
\cmidrule{4-5} 

\RowStyle{\bfseries}
Model & Arch. & \# Par. &  Words & Source & 
HuggingFace ID \\

\midrule

\gptbb &
\multirow{2}{*}{\gptb} & 
119 M & 100 M & 
\multirow{2}{*}{\shortstack{BabyLM, FineWeb-Edu,\\Cosmopedia}} &
ltg/gpt-bert-babylm-base
\\ 

\gptbs & 
&  
30 M & 10 M &
& 
ltg/gpt-bert-babylm-small 
\\
 
\midrule
\ltgbb &
\Block[c]{2-1}{\ltgb} &
99 M & 100 M & 
\multirow{2}{*}{BabyLM} & 
babylm/ltgbert-100m-2024
\\ 

\ltgbs &
&  
99 M & 10 M &
& 
babylm/ltgbert-10m-2024
\\

\midrule

\btimeb & 
\Block[c]{2-1}{\ltgb} &
98 M & 100 M & 
\Block[l]{2-1}{
BabyLM, TinyStories,\\\hspace{1em}GPT-Neo completions
} &
nikitastheo/BERTtime-Stories-100m-nucleus-1
\\ 

\btimes & 
& 
24 M & 10 M &
& 
nikitastheo/BERTtime-Stories-10m-nucleus-1-balanced
\\ 

\midrule
\eli & 
\rob &
44 M & 100 M & BabyLM + reddit ELI5  &
3van/RoBERTa\_100M\_ELI5\_ CurriculumMasking 
\\ 

\qecl & 
\rob &
43 M & 10 M & BabyLM - filtered QE &
jdebene/BabyLM2024
\\ 
\midrule

\robl  & 
\multirow{2}{*}{\rob} & 
304 M & $\sim$30 B & 
\multirow{2}{*}{\shortstack{Bookscorpus, English Wiki,\\ cc-news}} & 
FacebookAI/roberta-large
\\ 

\robb & 
& 
86 M & $\sim$30 B & 
& 
FacebookAI/roberta-base
\\

\midrule

\bertl & 
\multirow{2}{*}{\bbbb}& 
304 M & $\sim$3 B & 
\multirow{2}{*}{\shortstack{
Bookscorpus, English Wiki,\\openwebtext, stories 
}} &
google-bert/bert-large-cased \\

\bertb & 
& 
86 M & $\sim$3 B &  
& 
google-bert/bert-base-cased \\

\bottomrule

\end{NiceTabular}
\caption{Model Details}
\label{tab: models}
\end{table*}

\begin{table*}[t]
\begin{NiceTabular}{llp{1cm}p{1cm}lp{1cm}}
\toprule

\RowStyle{\bfseries}
& \multicolumn{5}{c}{Evals reported by BabyLM 2024 } \\ 
\cmidrule{2-6}
\RowStyle{\bfseries}
Model & BliMP & BLiMP Supplement & (Super) GLUE & EWoK & 
Macro \mbox{Avg} \\ 
\midrule

\gptbb & 86.1 & 76.8 & 81.5 & 58.4 & 75.7 \\ 
\ltgbb & 69.2 & 66.5 & 68.4 & 51.9 & 64 \\ 
\btimeb & 65.6 & 65 & 72.7 & 49.2 & 63.1 \\ 
\eli &  60.2 & 56.8 & 67.7 & 53 & 59.4 \\ 
\midrule

\gptbs & 81.2 & 69.4 & 76.5 & 54.6 & 70.4 \\ 
\btimes & 63.2 & 59.3 & 71.1 & 50.4 & 61 \\ 
\qecl & 61.9 & 58.3 & 64.4 & 50.8 & 58.8 \\ 
\ltgbs & 60.6 & 60.8 & 60.3 & 48.9 & 57.6 \\ 

\bottomrule

\end{NiceTabular}
\caption{BabyLM scores}
\label{tab: app: babylm score}
\end{table*}

\section{Methodological Details}

Experiments were run either on an M3 Macbook Pro or on a single Nvidia RTX A6000 cluster GPU.
Classification of idioms (MAGPIE) is the only experiment requiring any meaningful amount of computation and takes roughly 3 hours on a single RTX GPU.

\section{\babylm Supplement}
\label{app: babylms}

\autoref{tab: models} gives details for all models tested, and \autoref{tab: app: babylm score} reproduces each \babylm's scores that make up the macro average score reported in \autoref{tab: resuls main}.

The 2024 \babylmc is the second iteration of the \babylmc \cite{warstadt-etal-2023-findings}.
There were 18 models submitted to the strict track and 32 submitted to the strict small track. 
Of these models many support only autoregressive modeling and are thus not suited to testing using the bidirectional methods developed by \citeauthor{rozner2025} (in general, identifying constructions depends on bidirectional context). 
The best-performing model in both the strict and strict-small tracks was \gptb, a hybrid \mn{GPT} (autoregressive)/ \mn{BERT} (masked language model; MLM) architecture \citep{charpentier-samuel-2024-bert}, which was based on \ltgb \cite{samuel-etal-2023-trained}.
As \gptb can be run as an MLM it can be evaluated using \citeauthor{rozner2025}'s approach.

We select three additional models for both tracks to give eight total BabyLM models for evaluation.
For both tracks, we take the next top-performing MLMs (i.e., excluding purely autoregressive models), though we additionally exclude DeBERTa models because their tokenization scheme was not easily adapted to  work with the experiment pipeline. 
As this would have given us four \ltgb-style models for the strict track, we include a slightly worse-performing RoBERTa architecture for the final strict model.
Models are obtained from HuggingFace \cite{huggingface} at \url{https://huggingface.co/spaces/babylm/leaderboard-2024}, accessed March 18, 2025.

In the 2024 \babylmc, submissions were limited to a total \emph{quantity} of words but were not subject to any other limitations on training approaches. 
The best-performing \babylm model, \gptb, though limited to 100M or 10M words for the strict and strict-small tracks, still trained for many epochs over the same data \cite{hu-etal-2024-findings, wilcox2025bigger}. 
In the 2025 \babylmc, a limit is placed on the the number of times data can be seen \cite{charpentier2025babylm}.
For this study, as we are not treating PLMs as models of the learning process, but only as lower bounds on the linguistic knowledge that can be acquired from particular quantities of input, the particular details of training (e.g., more epochs) were not at issue.
Nonetheless, such divergences are interesting and would be relevant to any subsequent work that looks more closely at, e.g., the mechanisms or trajectories of acquisition.

\section{Evaluations}
\label{app: eval}

As the methods developed by \citet{rozner2025} are for singly-tokenized words, all words evaluated using either affinity method are singly-tokenized.
This constraint affects only the results for figurative vs. literal usage and NPN generalization studies, details of which we provide here.

\subsection{CEC vs. EAP/AAP}
\citeauthor{rozner2025} hypothesize that, as \ttw{so} is necessary in the CEC but not in the EAP/AAP (*\ttw{It was \textbf{very} big that it fell over}),
PLMs might identify this constraint in their output distribution. 
\citeauthor{rozner2025} find that global affinity in \robl on \ttw{so} in fact robustly distinguishes the CEC from EAP/AAP and that in multi-that sentences (e.g., \ttw{I was so happy that I won \textbf{that} I smiled}), the distribution for \ttw{so} is always more affected by the causal that.

The CEC dataset has 323 possible examples.
After cleaning and labeling, we have 292. 
Four are invalid, leaving us with a total of 22 EAP, 73 AAP, and 193 CEC (288 total).
\citeauthor{rozner2025} identify a couple mislabeled examples, and we use their corrected labels.

The accuracy we report for \robl is slightly lower than that reported by \citeauthor{rozner2025}. 
As we use a HuggingFace FastTokenizer, our evaluation included 15 examples that were omitted in \citeauthor{rozner2025}'s original study.
As these omitted examples had more awkward punctuation (tokenization issues were what led to their omission in the original study) that might have made them harder to correctly classify.

\paragraph{CEC multithat for so-that local affinity study}
We use the 31 entry multi-that dataset from \citeauthor{rozner2025}.

\subsection{Figurative vs. literal usages}
Here we provide two examples drawn from the MAGPIE dataset for \ttw{nuts and bolts} (same as \citeauthor{rozner2025}):
\\
\textbf{Literal usage}: They would include orders for routine raw materials such as steel stock; screws; \emph{nuts and bolts}; lubricants and fuel oil.
\\
\textbf{Figurative usage}: Jay comes from a different end of the spectrum to Dave Ambrose, but the two both like to talk \ttw{nuts and bolts}.

Magpie has 48,395 unique sentences with a total of 129,397 words used in potentially idiomatic expressions that are labeled as figurative or literal uses (average of 2.6 words per sentence/ example).
Of the 48,395 examples, we omit 3,944 \emph{sentences} that do not have $\geq 99\%$ confidence in annotation.
This leaves us with 44,451 sentences with a total of \mbox{119,401} words.
Among those 119,401 words, 2,016 have wrong offsets, giving us 117,385 words for the analysis. 
Of the 45,450 (117,385) sentences (words), 10,313 (23,484) are labeled as literal and 34,138 (95,917) are labeled as idiomatic.

For the result presented in the main paper, we omit from consideration any word or sentence that could not be processed for \emph{any} model (43,124 words).
In general, such processing failures are a result of differences in tokenization behavior:
the \citeauthor{rozner2025} methods are developed for singly-tokenized words and each \babylm has a different set of singly-tokenized words.
This allows us to make a fair comparison across all models.
This leaves us with 74,261 words for the analysis in the main paper.

Whereas \citeauthor{rozner2025} omit sentences with fewer than 10 words of context and words with less than four characters, we include all data.
The 69.2 score for \robl matches their corresponding result:
they report an AUC for the whole dataset of 0.71 with omission and 0.69 without omission. 
This correspondence suggests that even with this study's restriction to a common vocabulary over all models, the underlying trend in classification behavior is not substantially affected.

\subsection{Fixed slots in partially substantive constructions}
Examples of the six partially substantive constructions (from \citealt{bonial-tayyar-madabushi-2024-construction}), with fixed words italicized.
(These are the same as in \autoref{fig:cxns}.)
\begin{itemize}[label={},leftmargin=2pt]
    \item 
    \textbf{Causative-with:} She loaded the truck \ttw{with} books. 
    \item 
    \textbf{Comparative correlative:} \ttw{The} more \ttw{the} merrier. \\
    (In our analysis the two \ttw{the} words are considered as a single class.)
    \item 
    \textbf{Conative:} He kicked \ttw{at} the ball. 
    \item 
    \textbf{Let-alone:}  None of these arguments is particularly strong, \ttw{let alone} conclusive. 
    \item 
    \textbf{Much-less:}  He has not been put on trial, \ttw{much less} found guilty. 
    \item 
    \textbf{Way-manner:}  We made our \ttw{way} home. 
\end{itemize}

\subsection{Category constraint in the comparative correlative}

\begin{table}[t]
\centering
\begin{NiceTabular}{lrrr}

\toprule

Model & 50th \%\-ile & 80th \%\-ile & CC-score\\ 
\midrule
\gptbb & 2.2 & 3.1 & 99.7 \\ 
\ltgbb & 2.6 & 8.5 & 96.5 \\ 
\btimeb & 4.6 & 25.9 & 93.8 \\ 
\eli & 15.6 & 185.8 & 81.8 \\ 
\midrule
\gptbs & 2.6 & 6.2 & 96.4 \\ 
\btimes & 8.6 & 48.9 & 77.1 \\ 
\qecl & 59.6 & 818.0 & 49.1 \\ 
\ltgbs & 18.3 & 157.7 & 30.7 \\ 
\midrule
\robl & 2.1 & 2.6 & 99.9 \\ 
\robb & 2.2 & 3.4 & 99.5 \\ 
\bertl & 2.1 & 2.7 & 99.9 \\ 
\bertb & 2.5 & 5.4 & 98.2 \\ 

\bottomrule

\end{NiceTabular}
\caption{
CC: Average number of words (sorted by model likelihood) needed to get to nth \%-ile of output distribution for CC adj/adv slot. CC-score same as in \autoref{tab: resuls main} for comparison.
}
\label{tab: resuls cc supp}
\end{table}
We replicate the procedure of \citeauthor{rozner2025}:
Using the 54 CC examples from the CoGS dataset, 
we mask each comparative adjective/adverb, obtain the set of highest probability outputs at the masked position that sum to 85\% probability mass, and calculate a \emph{comparative score}: the percentage of this set that is a comparative adjective/adverb.

To calculate the percentage of the output distribution nucleus that is a comparative adj/adv, we order the outputs by probability and iterate through them until reaching a total probability mass of $p\geq0.85$ (a nucleus using 0.85).
\citeauthor{rozner2025} use a nucleus of 0.98, but since the BabyLM models have output distributions with much higher entropy than \robt (see \autoref{tab: resuls cc supp}), we consider a smaller nucleus to avoid summing probabilities over the whole vocabulary.

For each sampled word, we substitute it into the original sentence and use Spacy \cite{spacy} to check whether it is a comparative adverb or comparative adjective.
Whereas \citeauthor{rozner2025} use the transformer version of Spacy's tagger, we use the small, non-transformer model since the higher entropy distribution in the \babylms causes us to calculate for many more possible fills.
Given that the tagger module sees the whole sentence (and may already ``know'' the CC construction), it may be biased to label words as comparative even if they are not.
The final score is the proportion of the sample (the 85\% nucleus) that is a comparative adjective or adverb.

Of the 108 ($=54 \times 2$) candidate slots, across all twelve models, an average of 5.7 words cannot be processed (due to multi-tokenization).

\subsection{Generalization of the form of the NPN}
\label{app:npn}

\begin{table*}[t]
\centering
\begin{NiceTabular}{l *{12}{p{0.6cm}}}
\toprule

\RowStyle{\bfseries}
&
& \multicolumn{1}{c}{upon}  & &
& \multicolumn{1}{c}{after}  & & 
& \multicolumn{1}{c}{by}  & & 
& \multicolumn{1}{c}{to}  
\\

\cmidrule(lr){2-4}
\cmidrule(lr){5-7}
\cmidrule(lr){8-10}
\cmidrule(lr){11-13}

\RowStyle{\bfseries}
Model & 
All & Freq & Acc &
All & Freq & Acc &
All & Freq & Acc &
All & Freq & Acc
\\
\midrule
\gptbb & \colg{73}{73.4} & \colg{81}{81.3} & \colg{84}{83.7} & \colg{56}{55.6} & \colg{57}{57.2} & \colg{62}{61.6} & \colg{42}{42.3} & \colg{55}{55.1} & \colg{57}{56.8} & \colg{61}{60.5} & \colg{70}{70.2} & \colg{71}{70.6} \\ 
\ltgbb & \colg{61}{61.0} & \colg{66}{65.7} & \colg{68}{68.1} & \colg{39}{39.2} & \colg{42}{41.8} & \colg{45}{45.1} & \colg{50}{49.6} & \colg{63}{63.0} & \colg{65}{64.9} & \colg{31}{31.1} & \colg{38}{38.3} & \colg{38}{38.2} \\ 
\btimeb & \colg{39}{38.7} & \colg{43}{42.8} & \colg{46}{45.7} & \colg{22}{21.7} & \colg{20}{20.3} & \colg{23}{22.8} & \colg{34}{33.8} & \colg{43}{42.7} & \colg{45}{44.7} & \colg{8}{8.3} & \colg{11}{10.5} & \colg{10}{10.3} \\ 
\eli & \colg{0}{0.2} & \colg{0}{0.2} & \colg{0}{0.2} & \colg{0}{0.2} & \colg{0}{0.2} & \colg{0}{0.2} & \colg{1}{0.7} & \colg{0}{0.3} & \colg{1}{0.8} & \colg{0}{0.2} & \colg{0}{0.2} & \colg{0}{0.2} \\ 
\midrule
\gptbs & \colg{19}{19.3} & \colg{18}{17.7} & \colg{23}{23.2} & \colg{15}{14.8} & \colg{14}{13.8} & \colg{17}{17.3} & \colg{26}{26.1} & \colg{30}{29.7} & \colg{35}{35.4} & \colg{9}{9.2} & \colg{12}{11.8} & \colg{12}{11.7} \\ 
\btimes & \colg{26}{26.0} & \colg{30}{29.7} & \colg{30}{30.2} & \colg{15}{14.7} & \colg{13}{12.5} & \colg{15}{15.3} & \colg{34}{33.6} & \colg{45}{45.0} & \colg{46}{45.8} & \colg{3}{2.8} & \colg{3}{3.4} & \colg{3}{3.4} \\ 
\qecl & \colg{1}{0.6} & \colg{1}{0.5} & \colg{1}{0.5} & \colg{0}{0.3} & \colg{0}{0.1} & \colg{0}{0.1} & \colg{1}{1.3} & \colg{0}{0.2} & \colg{2}{2.2} & \colg{0}{0.2} & \colg{0}{0.2} & \colg{0}{0.2} \\ 
\ltgbs & \colg{0}{0.4} & \colg{0}{0.3} & \colg{0}{0.3} & \colg{0}{0.1} & \colg{0}{0.0} & \colg{0}{0.0} & \colg{1}{0.5} & \colg{0}{0.2} & \colg{1}{0.6} & \colg{0}{0.3} & \colg{0}{0.3} & \colg{0}{0.3} \\ 
\midrule
\robl & \colg{89}{88.6} & \colg{94}{94.4} & \colg{95}{95.2} & \colg{73}{72.6} & \colg{81}{80.7} & \colg{83}{82.7} & \colg{53}{52.8} & \colg{68}{68.3} & \colg{73}{72.5} & \colg{86}{86.3} & \colg{94}{94.1} & \colg{94}{94.2} \\ 
\robb & \colg{85}{84.5} & \colg{92}{91.5} & \colg{93}{92.8} & \colg{69}{69.0} & \colg{74}{73.5} & \colg{76}{76.4} & \colg{55}{54.9} & \colg{71}{70.9} & \colg{76}{75.6} & \colg{79}{79.4} & \colg{89}{88.9} & \colg{89}{89.0} \\ 
\bertl & \colg{90}{90.2} & \colg{95}{94.6} & \colg{95}{95.2} & \colg{73}{73.1} & \colg{81}{81.1} & \colg{83}{83.1} & \colg{53}{52.6} & \colg{66}{66.4} & \colg{69}{69.2} & \colg{80}{80.1} & \colg{87}{87.0} & \colg{87}{87.1} \\ 
\bertb & \colg{79}{79.1} & \colg{86}{85.8} & \colg{87}{87.2} & \colg{56}{55.6} & \colg{60}{60.0} & \colg{64}{63.6} & \colg{47}{46.6} & \colg{56}{55.8} & \colg{63}{63.2} & \colg{56}{55.9} & \colg{63}{63.1} & \colg{63}{63.3} \\ 

\bottomrule

\end{NiceTabular}
\caption{Results for NPN. 
All includes all 100 NPNs for each. 
Freq limits to acceptability $\geq4$ and further restricts to NPNs that occur 0 times in the \gptbb dataset.
Acc limits to NPNs with acceptability $\geq 4$.
Freq tends to have worse performance than Acc because it excludes NPNs that models were more likely to have seen, even if they are just as acceptable.
}
\label{tab: resuls npn}
\end{table*}

\paragraph{NPN dataset generation}
We follow \citeauthor{rozner2025}'s procedure in generating a new NPN dataset (below adapted from their Appendix).
We use GPT-4 via the OpenAI API, version gpt-4-0613, temperature 0.7, max tokens 100.
Total cost to produce 400 sentences is less than \$5.
We prompt as follows, where ``\{phrase\}'' is the particular targeted NPN (e.g., \ttw{day by day}):
\begin{quote}
An NPN construction is one like "day by day" or "face to face". 
It has a repeated singular noun with a preposition in the middle.
Other prepositions are also possible: "book upon book", "week over week", "year after year". 
Please use "\{phrase\}" in an NPN construction, 
placing "\{phrase\}" in the middle of the sentence. 
Make sure the sentence establishes a context in which the noun makes sense.
Please provide only the sentence in the response.
\end{quote}
We verify that each generation matches the desired form noun+prep+noun.

To obtain acceptability judgements, we randomly sort all sentences and the last author annotates with a score between 1 and 5, inclusive.

\paragraph{Results}
In the main text we reported only the average affinity score for NPNs using \ttw{upon} as preposition. 
In \autoref{tab: resuls npn} we report average scores for 
\begin{itemize}
\item 
all NPNs (all four prepositions: upon, after, by, to), 
\item 
all NPNs with acceptability $\geq4$ and filtered to those which occur zero times in the \gptbb training data (same as in \autoref{tab: resuls main})
\item 
all NPNs when filtered to acceptability $\geq4$.
\end{itemize}
(It is possible that some of the NPNs which have zero-frequency in the \gptbb training data have non-zero-frequency in the other models' data.)

When filtering to acceptable NPNs, we have upon: 72, after: 76, by: 64, to: 52.
Of these 264, 219 are seen zero times in \gptbb's training data.

Our results for \robl generally agree with the prior results of \citeauthor{rozner2025}: NPNs with \ttw{upon} are most well-generalized. 
Our results seem to show better generalization for NPNs using \ttw{to}, which could result from using the simpler common vocabulary across the \babylms.

Our overall NPN results show 
(i) gradient generalization of all NPNs, and
(ii) sensitivity to acceptability (more acceptable have higher affinity), agreeing with \citeauthor{rozner2025}'s results.
In general, removing NPNs that were in \gptbb's training data reduces average affinity scores. 
This makes sense given that there is overlap in the datasets for most of the models that were trained (all contain some part of the \babylm dataset). 
This implies an increase in affinity for NPNs that are observed in the training data.
This could be either an effect of memorization or it could reflect that NPNs in the training data have some underlying property that makes them more acceptable to the models, as measured by affinity.

\section{Use of AI Assistant}
ChatGPT was used to produce initial versions of some python matplot code.
Any code produced was subsequently adapted, reviewed, and/or modified.
ChatGPT was not used to write any part of this paper.

\end{document}